\documentclass[runningheads]{llncs}
\usepackage[T1]{fontenc}
\usepackage{newtxtext}       %
\usepackage[varvw]{newtxmath}       

\usepackage{graphicx}
\usepackage{booktabs}  
\usepackage{multirow}  
\providecommand{\Description}[1]{}  
\begin{document}
\title{ Beyond Binary Detection: Type-Aware Hallucination Analysis and Hierarchical Detection in Financial Question Answering }
\titlerunning{Type-Aware Hallucination Analysis in Financial QA}
%
%
\author{Xiongyu Chen\thanks{All authors contributed equally to this work.}\inst{1}\orcidID{0009-0006-6542-0147} \and
Sizhen Zhu\inst{2}\orcidID{0009-0000-5366-1257} \and
Zhuoer Wang\inst{3}\orcidID{0009-0005-3596-1391}}
\authorrunning{X. Chen et al.}
\institute{Carnegie Mellon University \email{chen22mail@gmail.com}\and
North Carolina State University \email{szhu13@alumni.ncsu.edu} \and University of Illinois Urbana-Champaign \email{zhuoerwang709@gmail.com}}
\maketitle              
\begin{abstract}
Large language models are increasingly used to answer questions over
financial filings, where an unfaithful answer can mislead investment,
compliance, or client decisions. Hallucination detectors are a common
safeguard, but they are usually evaluated with a single binary score that
reveals neither which kinds of errors analysesed nor why. This paper studies
hallucination detection on PHANTOM, a long-context benchmark built from SEC
filings, at the level of hallucination type. We first analyze 15,443 paired
hallucinated and faithful answers without any detector and find that
hallucinated answers are more than twice as long as faithful ones, and that
lexical grounding separates the two classes only because faithful answers are
bimodal. We then generate hallucinations in nine financially motivated
categories, confirm each with a separate verification pass, and diagnose an
LLM-based detector on 4,086 usable examples. Generation success varies from
66.3\% to 100\% across categories. The detector reaches 80.2\% binary
accuracy but only 44.7\% subtype accuracy, never assigns two of the nine
categories, and misses between 5.3\% and 46.8\% of hallucinations depending
on category; these misses are unrelated to input length but rise with how far
an answer departs from the correct one. Guided by this diagnosis, we propose a
two-stage hierarchical detector that first decides whether an answer is
hallucinated and then assigns a subtype, raising binary accuracy to 95.6\% and
subtype accuracy to 56.1\%. The results show that type-level evaluation
exposes detector weaknesses that aggregate metrics conceal.


\end{abstract}
\section{Introduction}\label{sec:intro}

Large language models (LLMs) are increasingly used to answer questions over
financial documents such as annual reports, current reports, and proxy
statements. In this setting, an answer that reads fluently but is unfaithful to
the source filing -- a misstated figure, a wrong reporting period, an invented
regulatory requirement -- can propagate into investment research, compliance
review, or client communication. Because such errors cannot be fully prevented
at generation time, a common safeguard is a separate hallucination detector,
typically an LLM prompted to judge whether a candidate answer is supported by
the source context, placed between the generator and the end user.

The quality of such detectors is usually summarised by a single binary accuracy
or F1 score on a benchmark. PHANTOM~\cite{ji2025phantom} is the most directly
relevant benchmark for financial documents: it pairs queries over SEC filings
with faithful and hallucinated answers, and expands each seed context to as many
as 30,000 tokens with the supporting evidence placed at different positions. Two
properties of this setting limit what an aggregate score can reveal. First, the
benchmark's hallucinations are dominated by one kind of error: its authors
classify 81.3\% of them as misrepresentations of the source, 17.9\% as
unverifiable information, and only 0.8\% as contradictions~\cite{ji2025phantom}.
Second, a binary score does not show which kinds of error a detector misses, or
whether its decisions rest on the evidence rather than on surface properties of
the answer. Both questions matter in finance, where a detector that reliably
catches altered numbers but misses a fabricated regulatory obligation gives a
false sense of coverage.

This paper studies hallucination detection on PHANTOM at the level of
hallucination type, in three steps. First, independently of any detector, we
analyse 15,443 pairs of hallucinated and faithful answers to identify surface
traits that separate the two classes (Section~\ref{sec:dataset-analysis}). Second, we
generate hallucinations in nine target categories, ranging from entity,
numerical, temporal, and regulatory hallucinations to reasoning errors,
unsupported inferences, instruction drift, and PHANTOM-style hallucinations,
confirm each generated
answer with a separate verification pass, and characterise where and why an
LLM-based detector succeeds or fails on the resulting 4,086 usable examples
(Section~\ref{sec:analysis}). Third, guided by this diagnosis, we propose a
two-stage hierarchical detector that first decides whether an answer is
hallucinated and only then assigns a subtype (Section~\ref{sec:modeling}).

The analysis surfaces several effects that an aggregate score hides.
Hallucinated answers in PHANTOM are on average more than twice as long as
faithful ones (76.5 vs.\ 35.0 words), a surface cue that a detector could
exploit without consulting the evidence. The generator itself is unreliable for
some categories, with landing rates ranging from 66.3\% to 100\%, so subtype
accuracy must be computed against verified rather than nominal category counts.
The baseline detector reaches 80.2\% binary accuracy but only 44.7\% subtype
accuracy, and it never assigns two of the nine categories, which points to a gap
in the label space exposed by its prompt rather than to graded difficulty. Its
miss rate ranges from 5.3\% to 46.8\% across categories and is unrelated to
context or query length, but it increases with how far a hallucinated answer
departs from the correct one: the detector catches subtle edits more reliably
than large rewrites. The two-stage detector raises binary accuracy to 95.6\%
and subtype accuracy to 56.1\%.

This paper makes four contributions. First, it provides a
detector-independent analysis of PHANTOM showing that answer length and lexical
grounding both carry label information, and that faithful answers are bimodal
in lexical grounding, so a single grounding threshold misclassifies loosely
paraphrased faithful answers. Second, it introduces a type-controlled
generate--verify--detect protocol over nine financial hallucination categories
that separates generator failure from detector failure. Third, it offers a
per-category diagnosis of an LLM-based detector that identifies a label-space
gap, rules out input length as a driver of missed detections, and shows that
larger departures from the correct answer are harder, not easier, to detect.
Finally, it proposes a two-stage hierarchical detector that substantially
improves both binary and subtype accuracy over the original detector.

The remainder of the paper is organised as follows.
Section~\ref{sec:related} reviews related work. Section~\ref{sec:dataset-analysis}
analyzes the dataset, Section~\ref{sec:analysis} diagnoses the baseline
detector, and Section~\ref{sec:modeling} presents and evaluates the two-stage
detector.

\section{Related Work}\label{sec:related}

Our work draws on four lines of research: hallucination benchmarks and
taxonomies in general domains, financial hallucination benchmarks, the
robustness of LLMs to long and noisy contexts, and the auditing of hallucination
detectors themselves. We review each in turn and situate our work relative to
them.

\subsubsection{General-domain benchmarks and taxonomies.}
HaluEval~\cite{li2023halueval} introduced a large benchmark of generated and
human-annotated hallucinations for question answering, knowledge-grounded
dialogue, and summarisation, but it is limited to short contexts and
general-domain text. HaluEval~2.0~\cite{li2024dawn} studies the sources and
mitigation of factuality hallucination and includes financial questions, but
without supporting documents, so it does not test faithfulness to a source.
Surveys organise hallucinations into factuality and faithfulness
errors~\cite{huang2025survey}, and several resources annotate finer-grained
types. RAGTruth~\cite{niu2024ragtruth} labels hallucinations in
retrieval-augmented generation (RAG) at the word level, distinguishing
conflicting from baseless content, and FAVA~\cite{mishra2024fava} defines a
taxonomy of entity, relation, sentence-level, invented, subjective, and
unverifiable errors and trains a span-level detector and editor. For RAG
evaluation more broadly, ARES~\cite{saadfalcon2024ares} trains lightweight
judges for context relevance, answer faithfulness, and answer relevance from a
small set of human annotations, and RAGEval~\cite{zhu2025rageval} generates
scenario-specific evaluation data, including financial scenarios.
HallusionBench~\cite{guan2024hallusionbench} addresses the analogous problem
for vision-language models. None of these resources targets long financial
filings or diagnoses a detector by hallucination type.

\subsubsection{Financial hallucination benchmarks.}
PHANTOM~\cite{ji2025phantom}, on which we build, constructs seed
query--context--answer triplets from chunks of roughly 500 tokens taken from
four SEC filing types (10-K, 8-K, 497K, and DEF~14A). Faithful and hallucinated
answers are generated by Llama-3.3-70B-Instruct, and a subset of seeds is
validated by human experts. Each seed is then expanded to contexts of 2K to 30K
tokens by surrounding the seed chunk with real filing text, with the evidence
placed at the beginning, middle, or end. The benchmark shows that detection
degrades as contexts grow, and that fine-tuning a 7B model on PHANTOM-style data
substantially improves it. Our work complements PHANTOM along an axis its design
does not vary, namely hallucination type: its own error breakdown is dominated
by misrepresentation (81.3\%), and its public review
discussion~\cite{phantom2025openreview} likewise noted limited coverage of other
modes such as entity swaps.

Other financial benchmarks approach the problem from different angles.
FAITH~\cite{zhang2025faith} frames intrinsic hallucination evaluation as
context-aware masked-span prediction over tables in S\&P~500 annual reports.
AlphaFin~\cite{li2024alphafin} and FinMTEB~\cite{tang2025finmteb} benchmark
financial analysis and financial text embeddings, respectively, and
OmniEval~\cite{wang2025omnieval} evaluates financial RAG systems along several
dimensions, including hallucination; none of them diagnoses detectors by
hallucination type. K-FinHallu~\cite{cho2026kfinhallu} is closest in spirit to
our work. It injects hallucinations into multi-turn Korean financial dialogues
through minimal edits that follow a hierarchical taxonomy (including
financial-term misunderstanding, number errors, and modifier changes), uses
hard-negative passages to create unanswerable cases, and reports both binary
detection and multi-class diagnosis, finding that even strong LLM detectors
struggle with fine-grained diagnosis. Our results reach a consistent conclusion
for English long-document question answering, where binary and subtype accuracy
diverge sharply. FinReflectKG--HalluBench~\cite{kumar2026finreflectkg} builds a
GraphRAG hallucination benchmark from 10-K filings and shows that adding noisy
knowledge-graph triplets lowers the Matthews correlation coefficient of most
detectors, by as much as 84\%, with LLM judges trusting structured signals over
textual evidence. Its error analysis identifies temporal misalignment,
hallucinated entities, incorrect numeric values, and cross-context
contamination, which overlap with several of our categories.
FinGround~\cite{guo2026finground} decomposes answers into atomic claims of six
financial types (e.g., numerical, temporal, and regulatory) and verifies each
type with a dedicated strategy. Like our two-stage detector, it exploits type
structure, but it is proposed as a new verification architecture rather than as
a diagnosis of where an existing detector fails.

\subsubsection{Robustness to long and noisy contexts.}
Outside hallucination detection, several studies show that both the length and
the content of a context affect LLM behaviour. Liu et al.~\cite{liu2024lost}
found that models use information at the beginning and end of long contexts
better than information in the middle. RGB~\cite{chen2024rgb} defines noise
robustness as a core RAG ability, varies the share of topically relevant
documents that do not contain the answer, and identifies concept confusion --
latching onto a similar but wrong concept in a noisy document -- as a distinct
failure mode. The Context Rot report~\cite{hong2025contextrot} documents
performance degradation across 18 LLMs as input length grows and as distractors
are added, while a recent reproducibility study~\cite{gabin2026lost} does not
always recover the U-shaped position effect but finds that sensitivity to
passage order and context size grows with the number of retrieved passages.
These studies measure generation, where a model must locate evidence and
compose an answer. A detector instead verifies an answer that is already given.
Liu et al.~\cite{liu2025longcontext} introduce a dataset for long-context
hallucination detection, and PHANTOM extends this setting to much longer
financial contexts. In contrast to the generation literature, we find that, on
verified hallucinations, the detector's miss rate is unrelated to context or
query length, whereas the type and magnitude of the edit matter.

\subsubsection{Auditing hallucination detectors.}
A further line of work asks whether detectors rely on the evidence they are
meant to check. Zollicoffer et al.~\cite{zollicoffer2026sanity} propose oracle
sanity checks that replace the final answer in a reasoning trace with the
correct one or remove answer-announcement steps, and show that several
long-form detectors rely on answer-level artefacts rather than on the reasoning
itself. Our dataset analysis raises an analogous concern for PHANTOM, in which
answer length alone separates the two classes. MedHal-Loc~\cite{chen2026medhalloc}
injects span-level errors of four types into medical statements and shows that
competitive detection F1 does not guarantee faithful localisation of the error.
The gap we observe between binary and subtype accuracy shows, similarly, that
recognising an error and characterising it are distinct abilities.
RT4CHART~\cite{yu2026retromorphic} decomposes RAG answers into claims that are
verified hierarchically, from local claims to the global answer, and finds
through re-annotation that existing benchmarks undercount hallucinations. Our
two-stage detector applies a hierarchy to the decision itself -- first whether
an answer is hallucinated, then which type -- rather than to the claims within
an answer.

In summary, prior work provides financial hallucination benchmarks,
fine-grained taxonomies, and tools for auditing detectors, but to our knowledge
these strands have not been combined into a type-level diagnosis of a detector
on financial long-context question answering. Our work provides such a
diagnosis, separates generator failure from detector failure through explicit
verification, and uses the resulting failure patterns to motivate a two-stage
detector.

\section{Dataset Analysis}
\label{sec:dataset-analysis}

We are using the \textsc{PhantomDataset} for our research. The dataset has 15{,}443 hallucination /
not-hallucination answer pairs spanning every filing type, context-token
counts, and context position in \textsc{PhantomDataset}, each pair sharing
the same query and source context. This helps us see which specific traits in an answer automatically make it look like a hallucination,
independent of any detector.

\subsection{Answer length is a consistent signal}
\label{sec:dataset-length}

Hallucinated answers are systematically longer than faithful ones
(Table~\ref{tab:dataset-length}): mean 76.5 words versus 35.0, with the
median gap nearly as large (67 vs.\ 25 words). The gap holds throughout the
distribution, not just at the mean -- the 25th percentile of hallucinated
answers (55 words) already exceeds the 75th percentile of not-hallucination
answers' lower half and sits well above the not-hallucination median. Faithful
answers also show far more variation at the short end (min 1 word vs.\ 28),
consistent with many of them being brief, direct extractions from the
context rather than elaborated responses.

\begin{table}[t]
\centering
\caption{Answer length (word count) by ground-truth label, over 15{,}443
pairs.}
\label{tab:dataset-length}
\begin{tabular}{lrrrrrrr}
\toprule
Label & Count & Mean & Std & Min & 25\% & 50\% & 75\% \\
\midrule
Hallucination      & 15{,}443 & 76.5 & 36.9 & 28 & 55 & 67 & 88 \\
Not hallucination  & 15{,}443 & 35.0 & 39.8 &  1 &  2 & 25 & 55 \\
\bottomrule
\end{tabular}
\end{table}

Whether the answer contains a digit at all is, by contrast, a weak signal:
71.1\% of hallucinated answers contain at least one number versus 66.9\% of
faithful answers, a 4.2 percentage-point gap that offers little practical
separation on its own.

\subsection{Lexical grounding cleanly separates the classes but hides a bimodal faithful-answer distribution}
\label{sec:dataset-grounding}

Context-grounding score -- the fraction of an answer's word $n$-grams that
also appear in the source context -- separates the two classes far more
cleanly than length or the presence of numbers
(Table~\ref{tab:dataset-grounding}). Median recall is 0.338 for
hallucinations versus 0.545 for faithful answers, a gap of roughly 0.21, and
the same pattern holds when grounding is computed at the sentence level and
aggregated by minimum or mean rather than over the whole answer at once, so
it is not an artefact of averaging across multi-sentence answers.

\begin{table}[t]
\centering
\caption{Answer grounding score (word 3-gram recall against the source
context) by ground-truth label, over 15{,}443 pairs.}
\label{tab:dataset-grounding}
\begin{tabular}{lrrrrrrr}
\toprule
Label & Count & Mean & Std & 25\% & 50\% & 75\% & Max \\
\midrule
Hallucination      & 15{,}443 & 0.356 & 0.156 & 0.241 & 0.338 & 0.453 & 0.923 \\
Not hallucination  & 15{,}443 & 0.521 & 0.376 & 0.150 & 0.545 & 0.908 & 1.000 \\
\bottomrule
\end{tabular}
\end{table}

The much larger standard deviation for not-hallucination answers (0.376 vs.\
0.156) indicates the two classes are not simply shifted versions of the same
shape. Not-hallucination answers are bimodal: a large cluster sits near
verbatim recall (75th percentile 0.908, max 1.000), consistent with these
being close quotes or paraphrases of the context, while a smaller cluster is
loosely paraphrased enough that its 25th percentile (0.150) actually falls
\emph{below} the hallucination class's 25th percentile (0.241). Hallucinated
answers, in contrast, occupy a narrow middle band throughout (IQR
0.241--0.453) and rarely reach either extreme -- they are neither
near-verbatim copies nor fully unmoored from the context, consistent with
hallucinations in this corpus being targeted edits or insertions into an
otherwise contextually grounded answer rather than fabrications built from
scratch.

The practical implication is that a single grounding-score threshold would
misclassify a meaningful share of the loosely paraphrased faithful answers as
hallucinations. Grounding score is a strong marginal signal, but on its own
it cannot distinguish "low overlap because paraphrased" from "low overlap
because fabricated"; a downstream detector would benefit from combining it
with a complementary signal such as answer length or a semantic
similarity/entailment score, rather than relying on lexical recall in
isolation. This is consistent with the weaker version of the same effect
observed once hallucinations are generated and passed through the detector
in Section~\ref{sec:analysis-grounding}.

\subsection{Summary}
\label{sec:dataset-summary}

Two findings carry forward into the detector analysis. First, hallucinated
answers in this corpus are on average twice as long as faithful ones, a
coarse but consistent surface signal, while the mere presence of a number in
the answer is not discriminative. Second, context-grounding score is the
strongest single lexical signal separating the classes (median gap
$\approx$0.21), but the faithful-answer class is bimodal -- split between
near-verbatim and heavily paraphrased answers -- so the signal is better
understood as "hallucinations occupy a narrow middle band" than as
"hallucinations are simply less grounded," and is best combined with a
complementary feature rather than thresholded alone.


\section{Two-Stage Hallucination Detector}
\label{sec:modeling}

As shown in the preceding analysis, hallucination detection performance varies substantially across hallucination types. To address this variation and improve detection across different types, we propose a hierarchical hallucination detector that organizes the detection process into multiple levels. This design aims to better capture type-specific characteristics and improve the benchmark detector’s overall accuracy.

\subsection{Methodology}
\label{sec:modeling-method}

Hallucination types can differ in both their defining features and the evidence needed to identify them. To help the detector distinguish whether an answer is hallucinated from what type of hallucination it contains, we reformulated the prompt as a two-stage decision process. 

To be more specific, it firstly determines whether a generated answer is hallucinated relative to the question, source context, and correct answer. Second, when the answer is flagged as hallucinated, it assigns one of eight targeted subtype labels. This design enforces a clean binary classification before subtype assignment and reduces ambiguity by requiring that subtype labels only be used when the first-stage decision is positive.

\subsubsection{Stage 1: Determine whether the answer is a hallucination.} The model assesses whether the generated answer is supported by the question and any provided context or reference answer. It assigns "Is Hallucination == False" only when the answer is sufficiently faithful to the available evidence.It assigns "Is Hallucination == True" when the answer contains unsupported, incorrect, or fabricated content. If the answer is not classified as a hallucination, subtype classification is not performed.

\subsubsection{Stage 2: Assign a subtype to hallucinated answers.} For answers classified as hallucinations in stage one, the model one label from the eight targeted subtype labels: entity hallucination, numerical hallucination, regulatory hallucination, temporal hallucination, reasoning hallucination, unsupported inference, reasoning errors, and instruction drift. The subtype decision is made using evidence-based matching rather than a waterfall or priority rule, so each subtype remains a valid candidate unless the evidence clearly supports a single best match.

This two-stage design is especially useful in evaluation settings where the binary task is substantially easier and more stable than the fine-grained classification task. It also makes model behavior easier to audit, because each prediction can be traced back to a clear first-pass decision and a second-pass subtype rationale.

\subsection{Performance Evaluation}
\label{sec:modeling-eval}
The final evaluation and assessment consider both overall classification performance and category-level metrics to provide a comprehensive view of the detector's strengths and limitations.

Two principal metrics are used to assess performance: binary accuracy and subtype accuracy. Binary accuracy measures the proportion of examples correctly classified as hallucinated or non-hallucinated. Subtype accuracy measures the proportion of correctly assigned hallucination subtypes among examples whose ground-truth labels indicate a hallucination.

\begin{table}[t]
\centering
\caption{Overall performance of the hallucination detector, measured by binary classification accuracy and subtype classification accuracy.}
\label{tab}
\begin{tabular}{lrrr}
\toprule
Metric & Result \\
\midrule
\texttt{Total evaluated samples} & 4,086 \\
\texttt{Binary classification accuracy} & 0.956 \\
\texttt{Subtype classification accuracy} & 0.561 \\
\bottomrule
\end{tabular}
\end{table}

The results indicate that the detector performs strongly on the binary classification task, achieving an accuracy of 95.57\%. This suggests that the model can reliably distinguish hallucinated answers from non-hallucinated answers across the evaluated dataset.

Subtype classification is more challenging, with an accuracy of 56.05\%. Although the model identifies hallucinations with high overall accuracy, distinguishing between fine-grained hallucination categories remains a more difficult task. This difference highlights the distinction between recognizing the presence of an error and determining its precise semantic or functional category.

\subsubsection{Binary Hallucination detector Performance}
Comparing to the benchmark detector, the enhanced two-stage Hallucination Detector performs better, which helps improve the accuracy ratio to 95.57\%.  

Especially, the binary detection task measures whether the model correctly identifies whether an answer is a hallucination or not. Comparing the original and enhanced detector runs shows a clear improvement across the active hallucination types.

\begin{table}[t]
\centering
\caption{Comparison of binary hallucination detection accuracy between the original and enhanced detectors across hallucination categories.}
\label{tab:original-vs-enhanced-binary-accuracy}
\begin{tabular}{lrr}
\toprule
Hallucination Type & Original Accuracy & Enhanced Accuracy \\
\midrule
\texttt{entity\_hallucination}     & 72.90\% & 92.29\% \\
\texttt{instruction\_drift}        & 88.60\% & 96.92\% \\
\texttt{numerical\_hallucination}  & 93.20\% & 98.90\% \\
\texttt{reasoning\_errors}         & 68.30\% & 94.71\% \\
\texttt{reasoning\_hallucination}  & 76.10\% & 96.04\% \\
\texttt{regulatory\_hallucination} & 89.70\% & 99.34\% \\
\texttt{temporal\_hallucination}   & 100.00\% & 98.68\% \\
\texttt{unsupported\_inference}    & 89.60\% & 94.27\% \\
\bottomrule
\end{tabular}
\end{table}

\begin{figure}
\includegraphics[width=\textwidth]{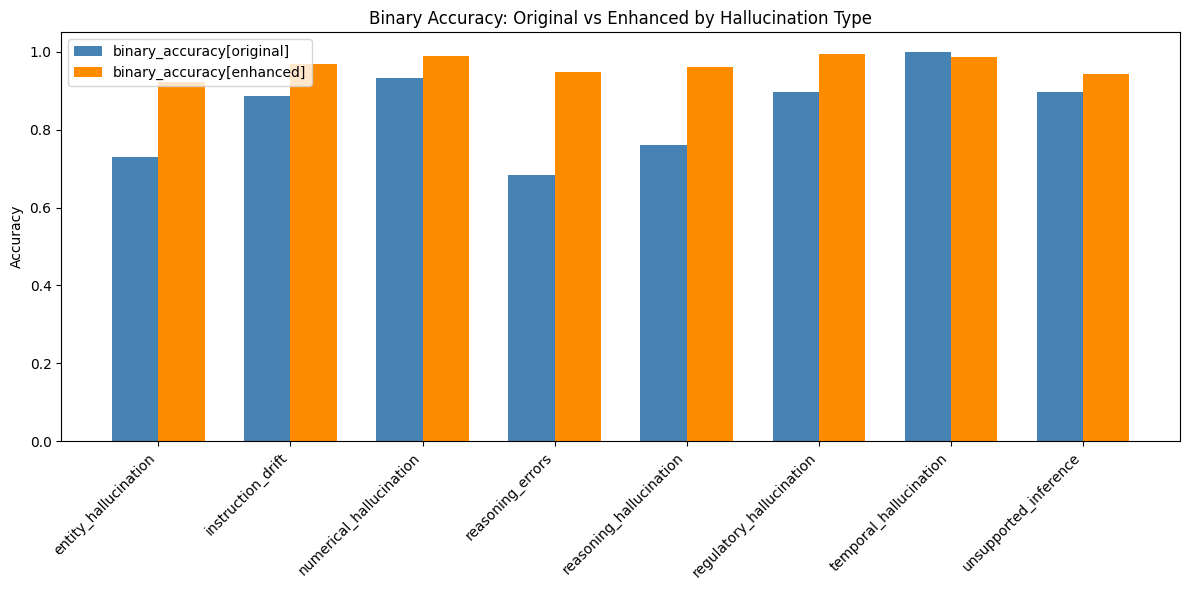}
\Description{In LaTeX you may insert your alternative texts into your manuscript using the \verb|\Description| command in a \verb|figure| environment}
\caption{A figure caption is always placed below the illustration.
Please note that short captions are centered, while long ones are
justified by the macro package automatically.} \label{pic}
\end{figure}

\noindent The enhanced detector shows stronger binary classification performance for nearly every category. The largest gains appear in:

\begin{itemize}
    \item \texttt{reasoning\_errors}: accuracy increased from 0.683 to 0.947.
    \item \texttt{reasoning\_hallucination}: accuracy increased from 0.761 to 0.960.
    \item \texttt{entity\_hallucination}: accuracy increased from 0.729 to 0.923.
    \item \texttt{numerical\_hallucination}: accuracy increased from 0.932 to 0.989.
\end{itemize}

These results indicate that the updated two-stage prompt and stronger decision rules improved the model's ability to distinguish genuine hallucinations from valid answers, particularly in subtle or reasoning-intensive cases. Overall, the enhanced detector provides more reliable binary classification, although further evaluation is needed to establish its robustness in ambiguous scenarios.

\subsubsection{Subtype Classification Performance}
On top of the hallucination detector, the enhanced prompt can also forecast the type of hallucination based on the feature of each subtype. The overall subtype accuracy of 56.05\% is calculated only on examples whose ground-truth labels indicate hallucinations. It measures the proportion of these examples for which the predicted subtype exactly matches the reference subtype.

\paragraph{Strong-Performing Categories.}

The results demonstrate substantial variation in performance across hallucination subtypes:

Regulatory hallucinations achieve the strongest performance, with an F1-score of 0.9633, precision of 0.9409, and recall of 0.9867. These results indicate that the detector identifies regulatory hallucinations with both high coverage and high reliability. One possible explanation is that regulatory errors often involve identifiable factual or rule-based discrepancies that are comparatively easier to distinguish from other error types.

Entity hallucinations also show strong performance, achieving an F1-score of 0.8373. The precision of 0.9055 indicates that predictions assigned to this category are generally reliable, while the recall of 0.7787 suggests that some entity-related errors are either missed or assigned to other categories.

Temporal hallucinations achieve an F1-score of 0.7789, with relatively balanced precision and recall. This indicates that the detector can identify many errors involving dates, time periods, or temporal relationships while maintaining a reasonable level of classification reliability.

Numerical hallucinations achieve an F1-score of 0.6832. Their recall of 0.8625 indicates that the detector identifies a substantial proportion of numerical errors. However, the lower precision of 0.5656 suggests that it frequently assigns the numerical hallucination label to examples belonging to other categories. This pattern indicates a tendency toward overprediction.

\paragraph{Weak-Performing Categories.}

The weakest results occur in categories that may require more complex semantic interpretation or distinctions between overlapping error types.

Instruction drift has an F1-score of 0.1556 and recall of 0.0859. Although its precision is relatively high at 0.8293, the detector predicts this category for only 41 examples despite a ground-truth support of 396. This indicates a pronounced underprediction problem: when the detector assigns this label, it is often correct, but it fails to identify most examples belonging to the category.

Reasoning errors achieve an F1-score of 0.2618, with precision of 0.7191 and recall of 0.1600. The detector appears to make conservative predictions for this category, identifying only a small proportion of the relevant examples. This may reflect difficulties distinguishing reasoning failures from unsupported inferences, reasoning hallucinations, or other semantically related categories.

Unsupported inference achieves an F1-score of 0.4404. Its recall of 0.7401 is substantially higher than its precision of 0.3134. The detector therefore captures many ground-truth unsupported inferences but also assigns this label to a large number of examples from other categories. The predicted count of 1,072, compared with a support of 454, reinforces the evidence of overprediction.

Reasoning hallucination achieves an F1-score of 0.4376, with precision of 0.4236 and recall of 0.4525. Both measures are relatively modest, suggesting that the detector has difficulty consistently distinguishing this category from other types of hallucination.

This two-stage design is especially useful in evaluation settings where the binary task is substantially easier and more stable than the fine-grained classification task. It also makes model behavior easier to audit, because each prediction can be traced back to a clear first-pass decision and a second-pass subtype rationale.

\begin{table}[t]
\centering
\caption{Hallucination subtype classification performance, including the number of ground-truth examples (support), predicted instances, true positives (TP), precision, recall, and F1-score for each category.}
\label{tab:hallucination-subtype-performance}
\begin{tabular}{lrrrrrr}
\toprule
Category & Support & Predicted & TP & Precision & Recall & F1-score \\
\midrule
\texttt{unsupported\_inference}    & 454 & 1072 & 336 & 0.3134 & 0.7401 & 0.4404 \\
\texttt{reasoning\_hallucination}  & 453 & 484  & 205 & 0.4236 & 0.4525 & 0.4376 \\
\texttt{regulatory\_hallucination} & 452 & 474  & 446 & 0.9409 & 0.9867 & 0.9633 \\
\texttt{temporal\_hallucination}   & 441 & 432  & 340 & 0.7870 & 0.7710 & 0.7789 \\
\texttt{numerical\_hallucination}  & 400 & 610  & 345 & 0.5656 & 0.8625 & 0.6832 \\
\texttt{reasoning\_errors}         & 400 & 89   & 64  & 0.7191 & 0.1600 & 0.2618 \\
\texttt{instruction\_drift}        & 396 & 41   & 34  & 0.8293 & 0.0859 & 0.1556 \\
\texttt{entity\_hallucination}     & 357 & 307  & 278 & 0.9055 & 0.7787 & 0.8373 \\
\bottomrule
\end{tabular}
\end{table}

\subsection{Conclusion}
\label{sec:modeling-conclusion}

The latest hallucination detector evaluation demonstrates strong binary classification performance, achieving 95.57\% accuracy across 4,086 evaluated examples. Subtype classification achieves 56.05\% accuracy among ground-truth hallucination examples, indicating that fine-grained error categorization remains a major area for improvement.

Category-level analysis shows particularly strong performance for regulatory hallucinations, entity hallucinations, and temporal hallucinations. However, the detector exhibits substantial weaknesses in instruction drift, reasoning errors, unsupported inference, and reasoning hallucination. The complete absence of predictions for 'default phantom' further highlights the need to verify label definitions, output constraints, and evaluation consistency.

Overall, the findings suggest that the detector has a strong foundation for binary hallucination identification but requires more targeted refinement to achieve reliable subtype classification. Future development should prioritize clearer subtype boundaries, correction of overprediction and underprediction patterns, structured classification rules, and systematic error analysis.

By addressing these issues and validating improvements on held-out data, the detector may achieve more consistent fine-grained classification while preserving its strong binary detection performance.

\section{Type-Aware Detection Analysis}
\label{sec:analysis}

The evaluation in Section 4 reports aggregate binary accuracy
(80.2\%) and subtype accuracy (44.7\%) for the detector. Here we go beyond
those top-line numbers to characterize \emph{where} and \emph{why} the
detector succeeds or fails, using the per-row generation, verification, and
detection outputs underlying that evaluation (4{,}085 usable rows, after
excluding one row for which the verification step itself failed).

\subsection{Generation fidelity varies sharply across target categories}
\label{sec:analysis-landing}

Each of the nine hallucination categories was attempted an equal 454 times,
but the generator does not reliably produce a genuine hallucination on every
attempt: a separate verification pass confirms whether the generated answer
actually is a hallucination, and, if so, whether it matches the intended
category. Table~\ref{tab:landing-rate} reports this \emph{landing rate}. Two
categories are markedly harder to induce than the rest: \texttt{default\_phantom}
lands only 66.3\% of the time and \texttt{entity\_hallucination} 78.8\%, while
five of the remaining seven categories land at $\geq$97\%. When an attempt
does land, it verifies as the intended category in $\geq$95\% of cases across
every category, so category drift at generation time is a minor effect
compared to outright generation failure.

This matters for interpreting subtype accuracy: because \texttt{default\_phantom}
and \texttt{entity\_hallucination} contribute far fewer verified examples than
their nominal 454-per-category design would suggest (301 and 357,
respectively), any accuracy computed against the nominal design count
conflates generator failure with detector failure. The per-category support
figures in Section 4.2 are computed against the verified
counts, which avoids this, but it is worth stating explicitly since it is not
obvious from the aggregate numbers alone.

\begin{table}[t]
\centering
\caption{Landing rate by attempted hallucination category: the fraction of
generation attempts that verification confirmed as genuine hallucinations,
and, among those, the fraction that matched the intended category.}
\label{tab:landing-rate}
\begin{tabular}{lrrr}
\toprule
Category & Attempted & Landed & Landing rate \\
\midrule
\texttt{default\_phantom}          & 454 & 301 & 66.3\% \\
\texttt{entity\_hallucination}     & 453 & 357 & 78.8\% \\
\texttt{instruction\_drift}        & 454 & 396 & 87.2\% \\
\texttt{numerical\_hallucination}  & 454 & 400 & 88.1\% \\
\texttt{reasoning\_errors}         & 454 & 400 & 88.1\% \\
\texttt{temporal\_hallucination}   & 454 & 441 & 97.1\% \\
\texttt{regulatory\_hallucination} & 454 & 452 & 99.6\% \\
\texttt{reasoning\_hallucination}  & 454 & 453 & 99.8\% \\
\texttt{unsupported\_inference}    & 454 & 454 & 100.0\% \\
\bottomrule
\end{tabular}
\end{table}

\subsection{Two categories are structurally absent from the detector's output}
\label{sec:analysis-taxonomy}

\begin{figure}[t]
\centering
\includegraphics[width=0.9\linewidth]{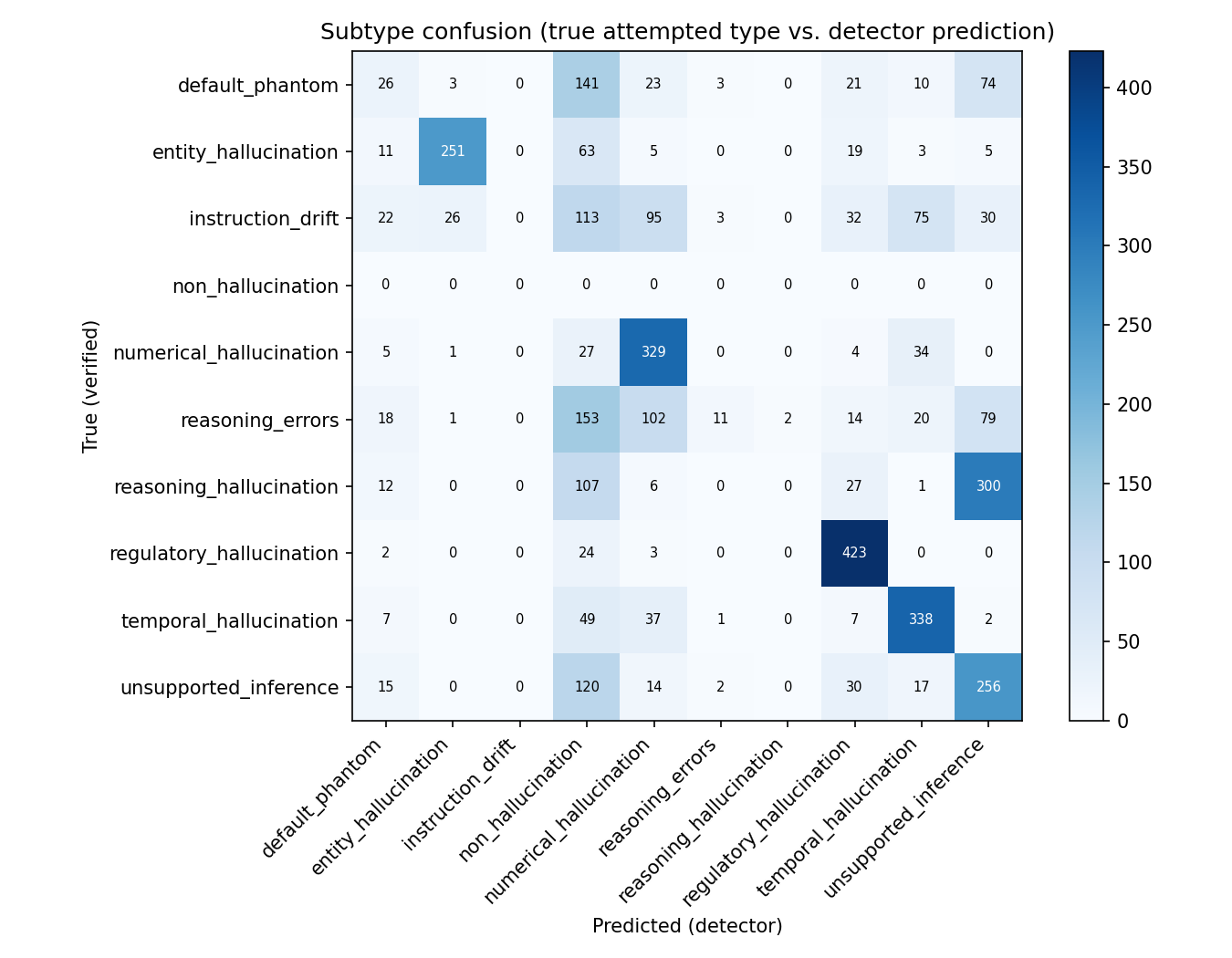}
\caption{Subtype confusion matrix: true attempted category (rows) vs.\
detector-predicted category (columns), restricted to verified hallucinations.
The all-zero diagonal cells for \texttt{instruction\_drift} and
\texttt{reasoning\_hallucination} are visible as the two rows with no
diagonal mass.}
\label{fig:confusion}
\end{figure}

The full subtype confusion matrix (true attempted category vs.\ detector
prediction, restricted to the 3{,}654 verified hallucinations;
Figure~\ref{fig:confusion}) reveals a failure mode distinct from ordinary
misclassification. For \texttt{instruction\_drift} and
\texttt{reasoning\_hallucination}, the diagonal is exactly zero: the detector
never once assigns the correct label, and across the entire 4{,}085-row
evaluation set it emits \texttt{instruction\_drift} zero times and
\texttt{reasoning\_hallucination} only twice (both incorrect). This is not a
matter of these categories being merely difficult -- every other category is
predicted correctly at least 60\% of the time when it lands-- but of the detector apparently never treating them as
selectable outputs at all, which points to a mismatch between the
generator's category taxonomy and the label space the detector's prompt
actually exposes, rather than a graded detection difficulty.

Table~\ref{tab:taxonomy-gap} shows where these two categories' verified hallucinations get reassigned instead. \texttt{reasoning\_hallucination}
mostly reassigned into \texttt{unsupported\_inference} (66.2\%), a plausible semantic neighbor. \texttt{instruction\_drift} disperses more
broadly, most often into \texttt{non\_hallucination} (28.5\%) and
\texttt{numerical\_hallucination} (24.0\%) -- the latter a much less obvious
neighbor, consistent with the detector reaching for whichever label is
nominally closest rather than recognizing the category at all. Because this
gap plausibly reflects the detector's label schema rather than a modeling
weakness, we consider revisiting the detector's prompt to confirm both
categories are actually enumerated as options to be the single
highest-leverage fix available, ahead of any feature-level improvement
discussed below.

\begin{table}[t]
\centering
\caption{Where verified hallucinations of the two "missing" categories get
reassigned by the detector (share of that category's landed hallucinations;
top three destinations shown).}
\label{tab:taxonomy-gap}
\begin{tabular}{llr}
\toprule
True category & Predicted as & Share \\
\midrule
\multirow{3}{*}{\texttt{instruction\_drift} ($n{=}396$)}
  & \texttt{non\_hallucination}       & 28.5\% \\
  & \texttt{numerical\_hallucination} & 24.0\% \\
  & \texttt{temporal\_hallucination}  & 18.9\% \\
\midrule
\multirow{3}{*}{\texttt{reasoning\_hallucination} ($n{=}453$)}
  & \texttt{unsupported\_inference} & 66.2\% \\
  & \texttt{non\_hallucination}     & 23.6\% \\
  & \texttt{regulatory\_hallucination} & 6.0\% \\
\bottomrule
\end{tabular}
\end{table}

\subsection{Binary miss rate is category-dependent but not explained by length}
\label{sec:analysis-missrate}

Besides subtype, the detector's binary miss rate -- the fraction of hallucinations it labels \texttt{non\_hallucination} outright --
ranges from 5.3\% for \texttt{regulatory\_hallucination} to 46.8\% for
\texttt{default\_phantom} (Table~\ref{tab:miss-rate}). This is a different
cut than category-conditional recall reported in Section 4.3,
since it ignores subtype correctness and asks only whether the detector
flagged the row as a hallucination at all.

A natural confound to rule out is input size: perhaps longer contexts or
questions simply give the detector more surface area to miss something in.
We do not find evidence for this. Binning verified hallucinations into
context-length and query-length quartiles produces a flat miss rate (19--24\%
in every bin), and a Kruskal-Wallis test confirms neither context length nor query length differs
significantly even across the nine hallucination categories themselves --
in contrast to every other feature we tested (Table~\ref{tab:kruskal}), all
of which differ across categories at $p<0.001$. Input length is simply not
informative here, for either the detector or the category taxonomy.

What does show a length-independent difficulty gradient is the underlying
\emph{query}: grouping the 454 distinct queries (each reused across all nine
attempted categories) by their detector miss rate gives a mean of 21.9\%
(s.d.\ 17.8\%) but a wide spread -- 80 queries (17.6\%) have every one of
their landed hallucination attempts caught, while 4 queries have every
landed attempt missed regardless of category. This suggests a
source-document- or topic-level difficulty factor independent of
hallucination category, which the present feature set does not capture and
which we leave to future work.

\begin{table}[t]
\centering
\caption{Detector miss rate by category (share of verified hallucinations
labeled \texttt{non\_hallucination}), and the single feature most predictive
of a miss within that category (Mann-Whitney $U$ test, Bonferroni-corrected
$\alpha = 0.01/9 \approx 0.0011$ across the nine candidate features).
$\dagger$ denotes significance surviving correction.}
\label{tab:miss-rate}
\begin{tabular}{lrll}
\toprule
Category & Miss rate & Top predictor of a miss & $p$ \\
\midrule
\texttt{default\_phantom}          & 46.8\% & context grounding (lower)          & 0.045 \\
\texttt{reasoning\_errors}         & 38.2\% & edit distance from correct (higher)$\dagger$ & $<$0.001 \\
\texttt{instruction\_drift}        & 28.5\% & edit distance from correct (higher)$\dagger$ & $<$0.001 \\
\texttt{unsupported\_inference}    & 26.4\% & context grounding (lower)$\dagger$ & $<$0.001 \\
\texttt{reasoning\_hallucination}  & 23.6\% & correct-answer overlap (lower)$\dagger$ & $<$0.001 \\
\texttt{entity\_hallucination}     & 17.6\% & number overlap (lower)             & 0.028 \\
\texttt{temporal\_hallucination}   & 11.1\% & edit distance from correct (higher)& 0.003 \\
\texttt{numerical\_hallucination}  &  6.8\% & edit distance from correct (higher)& 0.001 \\
\texttt{regulatory\_hallucination} &  5.3\% & correct-answer overlap (lower)$\dagger$ & $<$0.001 \\
\bottomrule
\end{tabular}
\end{table}

\subsection{What separates caught from missed hallucinations differs by category}
\label{sec:analysis-features}

Beyond miss rate, we characterized each category's generated answers along
four dimensions: length, lexical grounding in the source context ($n$-gram
recall), word-level edit distance from the correct answer, and word-overlap
(Jaccard) similarity to the correct answer. A Kruskal-Wallis test confirms
the nine categories are genuinely distinguishable on every one of these
dimensions ($p<0.001$; Table~\ref{tab:kruskal}), with edit distance from the
correct answer the most discriminating single feature. Qualitatively,
\texttt{default\_phantom} and \texttt{reasoning\_hallucination} manifest as
long insertions with low overlap to the correct answer (mean length 87--89
words, Jaccard 0.30--0.42), while \texttt{numerical\_hallucination} and
\texttt{entity\_hallucination} are short, surgical edits with high overlap
(47--52 words, Jaccard 0.58--0.60) -- consistent with swapping a single
number or named entity while leaving the rest of the answer untouched.

\begin{table}[t]
\centering
\caption{Kruskal-Wallis test for whether each feature's distribution differs
across the nine attempted hallucination categories.}
\label{tab:kruskal}
\begin{tabular}{lrl}
\toprule
Feature & $H$ & $p$ \\
\midrule
Edit distance from correct answer & 1633.2 & $<$0.001 \\
Entity-token overlap w/ correct    & 1336.1 & $<$0.001 \\
Number-token overlap w/ correct    &  997.0 & $<$0.001 \\
Jaccard overlap w/ correct answer  &  545.2 & $<$0.001 \\
Generated answer length            &  478.0 & $<$0.001 \\
Sentence-count delta               &  473.3 & $<$0.001 \\
Context grounding (n-gram recall)  &  268.4 & $<$0.001 \\
Query length                       &    2.4 & 0.967 \\
Context length                     &    5.1 & 0.748 \\
\bottomrule
\end{tabular}
\end{table}

\begin{figure}[t]
\centering
\includegraphics[width=\linewidth]{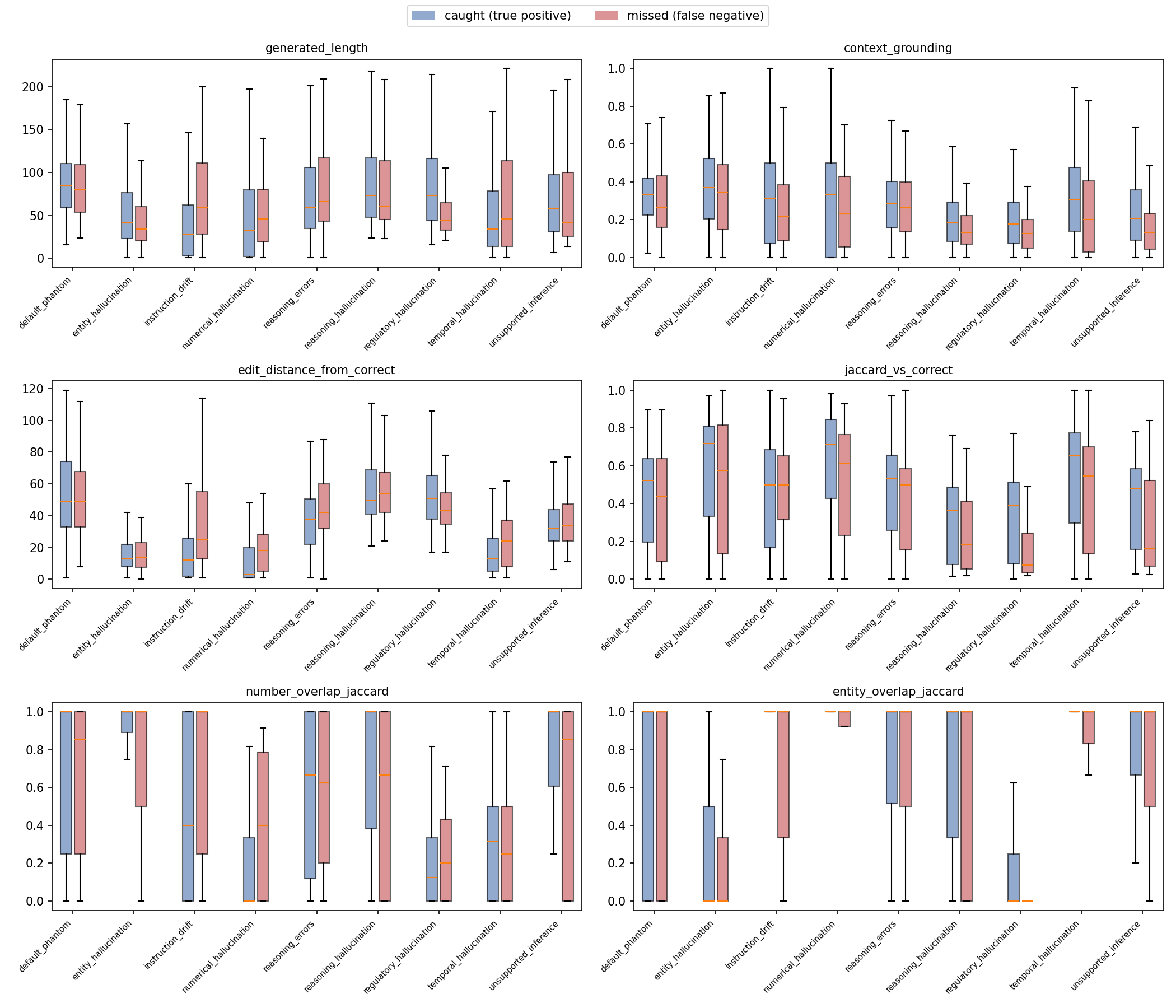}
\caption{Distribution of six candidate features, split by category and by
whether the detector caught (blue) or missed (red) the verified
hallucination.}
\label{fig:boxplots}
\end{figure}

Repeating the caught-vs.-missed comparison of Section~\ref{sec:analysis-missrate}
\emph{within} each category (rightmost columns of Table~\ref{tab:miss-rate};
full results and per-category boxplots in Figure~\ref{fig:boxplots}) shows one
consistent pattern rather than nine unrelated ones. Whichever of edit
distance, correct-answer overlap, or context grounding reaches significance
for a given category, the direction is always the same: \emph{greater}
divergence from the correct answer or from the source context predicts a
\emph{miss}, not a catch. This holds via edit distance for
\texttt{instruction\_drift}, \texttt{reasoning\_errors},
\texttt{numerical\_hallucination}, and \texttt{temporal\_hallucination}
(larger rewrites are harder to catch), and via correct-answer or
context overlap for \texttt{reasoning\_hallucination},
\texttt{regulatory\_hallucination}, and \texttt{unsupported\_inference}
(lower overlap is harder to catch). Both directions run counter to the
intuition that a larger, more obviously fabricated rewrite should be easier
to flag; instead the detector is comparatively good at catching close,
subtle edits and comparatively poor at catching answers that drift furthest
from the ground truth. A pooled logistic regression across all categories
recovers the same direction on both features, so this is not an artifact of
the per-category split -- the per-category view mainly clarifies that the
pooled effect is carried by seven of the nine categories rather than being
universal, since \texttt{default\_phantom} and \texttt{entity\_hallucination}
show no significant relationship on any feature (discussed below).

One feature does flip sign across categories: number-token overlap with the
correct answer. For \texttt{instruction\_drift}, retaining more of the
correct answer's numbers predicts a \emph{miss} (mean overlap 0.685 for
missed vs.\ 0.482 for caught, $p<0.001$) -- consistent with the dominant
pattern above, since the numeric content is the part of the answer that
stayed close to ground truth. For \texttt{unsupported\_inference}, the
opposite holds: retaining more of the correct answer's numbers predicts
being \emph{caught} (0.784 vs.\ 0.608, $p<0.001$). This is plausible given how the two categories are constructed: \texttt{instruction\_drift} alters
existing content, so preserved numbers mask the drift, whereas
\texttt{unsupported\_inference} appends new, unsupported claims onto an
otherwise-intact answer, so an addition that leaves every original number
untouched is comparatively easy for the detector to isolate as an
insertion.

Two categories stand out for the opposite reason: for
\texttt{default\_phantom} and \texttt{entity\_hallucination}, no feature we
tested survives Bonferroni correction as a significant predictor of a miss
within that category. Whatever distinguishes the hallucinations the detector
catches from the ones it misses in these two categories is apparently not
captured by surface length, lexical grounding, or similarity to the correct
answer, and likely requires closer qualitative reading of individual missed
cases rather than further feature engineering.

\subsection{Lexical grounding alone is a much weaker signal on this dataset than on the underlying corpus}
\label{sec:analysis-grounding}

Prior work on the underlying document corpus (i.e., comparing raw
hallucinated vs.\ non-hallucinated answers directly, without a detector in
the loop) found that context-grounding score -- the fraction of an answer's
word $n$-grams that also appear in the source context -- cleanly separated
the two classes (median recall 0.338 for hallucinations vs.\ 0.545 for
faithful answers, a gap of roughly 0.21). Repeating that same comparison on
the generated-and-verified answers analyzed here yields a substantially
smaller gap: mean grounding of 0.271 for verified hallucinations vs.\ 0.321
for verified non-hallucinations, less than a third of the original
separation, though still in the same direction. Because this generation
pipeline produces hallucinations as targeted edits to an otherwise-correct
answer (Section~\ref{sec:analysis-features}) rather than loosely paraphrased
fabrications, the hallucinated answers remain substantially grounded in the
context by construction. This is consistent with the detector's imperfect
80.2\% binary accuracy: a detector relying substantially on lexical overlap
with the context would be expected to perform markedly worse here than on
the original corpus, and a purely lexical grounding signal is unlikely to
close much of the remaining gap without a complementary, more semantic
signal (e.g., natural language inference against the specific claim being
checked).

\subsection{Summary}
\label{sec:analysis-summary}

Four findings emerge. First, the generator itself fails to produce a genuine
hallucination for 1/3  \texttt{default\_phantom} attempts and a fifth
of \texttt{entity\_hallucination} attempts, so subtype accuracy must be
computed against verified, not nominal, category counts. Second, two of the
nine categories (\texttt{instruction\_drift}, \texttt{reasoning\_hallucination})
are essentially never produced as detector outputs at all -- a likely
taxonomy gap in the detector's own prompt, and the highest-leverage fix
identified in this analysis. Third, detection difficulty is driven by each
category's degree of divergence from the correct answer, and consistently in
the counter-intuitive direction -- larger, more visibly divergent rewrites
are \emph{harder} for the detector to catch than close, subtle edits, not
easier -- rather than by input length, which has no measurable effect.
Fourth, the lexical
context-grounding signal that separates hallucinated from faithful answers
cleanly in the raw corpus is far weaker once hallucinations are generated as
targeted, well-grounded edits, motivating a semantic rather than purely
lexical detection signal going forward.

\section{Conclusion}
This paper investigated hallucination detection in financial long-context question answering context, with a particular focus on understanding how detection performance varies across different hallucination types. Using the PHANTOM benchmark and a hallucination generation and verification framework, we examined both the characteristics of hallucinated answers and the limitations of existing LLM-based detection approaches. Based on these findings, we proposed a two-stage hierarchical detector to improve hallucination identification and fine-grained classification.

Our analysis revealed several important limitations of conventional hallucination detection. First, although answer length and lexical grounding provide useful signals for distinguishing hallucinated from faithful answers, these surface-level characteristics are insufficient and less explainable for reliable detection, particularly when hallucinations involve subtle modifications or contextually plausible statements. Second, detection performance varies substantially between hallucination categories, revealing weaknesses that aggregate binary accuracy cannot adequately capture. In particular, our experiments showed that hallucinations involving larger deviations from correct answers were often more difficult to detect, while the length of the input context was not a significant predictor of the detection failure in the evaluated setting. These findings highlight the importance of evaluating hallucination detectors beyond conventional binary classification metrics.

To address the identified limitations, we introduced a two-stage hierarchical detection approach that separates binary hallucination identification from subtype classification. Experimental results demonstrated substantial improvements over the original detector, increasing binary classification accuracy from 80.2\% to 95.6\% and subtype classification accuracy from 44.7\% to 56.1\%. The enhanced detector achieved particularly strong performance in identifying regulatory, entity, and temporal hallucinations. Nevertheless, classification performance remained uneven across categories, especially for instruction drift, reasoning errors, and unsupported inference, indicating that fine-grained semantic distinctions remain challenging.

These findings suggest that hierarchical detection provides a promising direction for improving the reliability and interpretability of LLM-based hallucination detection in financial applications. However, the remaining subtype classification errors highlight the need for more precise category definitions and stronger evidence-based reasoning. Future research could explore claim-level verification, semantic entailment techniques, and category-specific detection strategies, together with evaluation across additional financial documents and model architectures.

Overall, this study demonstrates that reliable hallucination detection requires not only identifying whether an answer is incorrect or unsupported, but also understanding the nature of the underlying error. By combining type-level diagnostic analysis with hierarchical classification, our work provides a foundation for developing more robust and transparent hallucination detection systeHallucinationial question answering.

%
%
%

\end{document}